\documentclass[10pt,twocolumn,twoside]{IEEEtran}
\IEEEoverridecommandlockouts
\usepackage{epsfig}
\usepackage{graphicx}
\usepackage{amsmath}
\usepackage{amssymb}
\usepackage{multirow}
\usepackage{subfigure}
\usepackage{algorithm}
\usepackage{algorithmic}
\usepackage{subfigure}
\usepackage{color}
\usepackage{epstopdf}

\usepackage{hyperref}
\graphicspath{{figures/}}  

\begin{document}

%%%%%%%%% TITLE
\title{Weakly Supervised PatchNets: Describing and Aggregating Local Patches for Scene Recognition}

\author{Zhe~Wang,~Limin~Wang,~Yali~Wang,~Bowen~Zhang,~and~Yu~Qiao,~\IEEEmembership{Senior~Member,~IEEE}% <-this % stops a space

\thanks{This work was supported in part by National Key Research and Development Program of China (2016YFC1400704), National Natural Science Foundation of China (U1613211, 61633021, 61502470), Shenzhen Research Program (JCYJ20160229193541167), and External Cooperation Program of BIC, Chinese Academy of Sciences, Grant 172644KYSB20160033.}
\thanks{Z. Wang was with the Shenzhen Institutes of Advanced Technology, Chinese Academy of Sciences, Shenzhen, China, and is with the Computational Vision Group, University of California, Irvine, CA, USA (buptwangzhe2012@gmail.com).}
\thanks{L. Wang is with the Computer Vision Laboratory, ETH Zurich, Zurich, Switzerland (07wanglimin@gmail.com).} 
\thanks{Y. Wang is with the Shenzhen Institutes of Advanced Technology, Chinese Academy of Sciences, Shenzhen, China (yl.wang@siat.ac.cn).}
\thanks{B. Zhang is with the Shenzhen Institutes of Advanced Technology, Chinese Academy of Sciences, Shenzhen, China. He is also with Tongji University, Shanghai, China (1023zhangbowen@tongji.edu.cn).}
\thanks{Y. Qiao is with the Shenzhen Institutes of Advanced Technology, Chinese Academy of Sciences, Shenzhen, China. He is also with the Chinese University of Hong Kong, Hong Kong (yu.qiao@siat.ac.cn).}
}

\markboth{IEEE Transactions on Image Processing}%
{Weakly Supervised PatchNets: Describing and Aggregating Local Patches for Scene Recognition}

\maketitle

%%%%%%%%% ABSTRACT
\begin{abstract}
Traditional feature encoding scheme (e.g., Fisher vector) with local descriptors (e.g., SIFT) and recent convolutional neural networks (CNNs) are two classes of successful methods for image recognition. In this paper, we propose a hybrid representation, which leverages the discriminative capacity of CNNs and the simplicity of descriptor encoding schema for image recognition, with a focus on scene recognition. To this end, we make three main contributions from the following aspects. First, we propose a patch-level and end-to-end architecture to model the appearance of local patches, called {\em PatchNet}. PatchNet is essentially a customized network  trained in a weakly supervised manner, which uses the image-level supervision to guide the patch-level feature extraction. Second, we present a hybrid visual representation, called {\em VSAD}, by utilizing the robust feature representations of PatchNet to describe local patches and exploiting the semantic probabilities of PatchNet to aggregate these local patches into a global representation. Third, based on the proposed VSAD representation, we propose a new state-of-the-art scene recognition approach, which achieves an excellent performance on two standard benchmarks: MIT Indoor67 (86.2\%) and SUN397 (73.0\%).
\end{abstract}


%%%%%%%%% BODY TEXT
\IEEEpeerreviewmaketitle

\section{Introduction}
\label{Sec:Introduction}

Image recognition is an important and fundamental problem in computer vision research~\cite{KrizhevskySH12,HeZRS16,ZhouLXTO14,ShenLH15,XiongZLT15,WangWDQ15,WangWQG16}. Successful recognition methods have to extract effective visual representations to deal with large intra-class variations caused by scale changes, different viewpoints, background clutter, and so on. Over the past decades, many efforts have been devoted to extracting good representations from images, and these representations may be roughly categorized into two types, namely {\em hand-crafted} representations and {\em deeply-learned} representations.

In the conventional image recognition approaches, hand-crafted representation is very popular due to its simplicity and low computational cost. Normally, traditional image recognition pipeline is composed of feature extraction, feature encoding (aggregating), and classifier training. In feature extraction module, local features, such as SIFT~\cite{Lowe04}, HOG~\cite{DalalT05}, and SURF~\cite{BayETG08}, are extracted from densely-sampled image patches. These local features are carefully designed to be invariant to local transformation yet able to capture discriminative information. Then, these local features are aggregated with a encoding module, like Bag of Visual Words (BoVW)~\cite{csurka2004,SivicZ03}, Sparse coding~\cite{YangYGH09}, Vector of Locally Aggregated Descriptor (VLAD)~\cite{JegouPDSPS12}, and Fisher vector (FV)~\cite{PerronninSM10,SanchezPMV13}. Among these encoding methods, Fisher Vector and VLAD can achieve good recognition performance with a shallow classifier (e.g., linear SVM).

Recently, Convolutional Neural Networks (CNNs)~\cite{lecun-98} have made remarkable progress on image recognition since the ImageNet Large Scale Visual Recognition Challenge (ILSVRC) 2012~\cite{RussakovskyDSKS15}. These deep CNN models directly learn discriminative visual representations from raw images in an end-to-end manner. Owing to the available large scale labeled datasets (e.g., ImageNet~\cite{DengDSLL009}, Places~\cite{ZhouLXTO14}) and powerful computing resources (e.g., GPUs and parallel computing cluster), several successful deep architectures have been developed to advance the state of the art of image recognition, including AlexNet~\cite{KrizhevskySH12}, VGGNet~\cite{SimonyanZ14a}, GoogLeNet~\cite{SzegedyLJSRAEVR15}, and ResNet~\cite{HeZRS16}. Compared with conventional hand-crafted representations, CNNs are equipped with rich modeling power and capable of learning more abstractive and robust visual representations. However, the training of CNNs requires large number of well-labeled samples and long training time even with GPUs. In addition, CNNs are often treated as black boxes for image recognition, and it is still hard to well understand these deeply-learned representations. 

In this paper we aim to present a hybrid visual representation for image recognition, which shares the merits of hand-crafted representation (e.g., simplicity and interpretability) and deeply-learned representation (e.g., robustness and effectiveness). Specifically, we first propose a patch-level architecture to model the visual appearance of a small region, called as PatchNet, which is trained to maximize the performance of image-level classification. This weakly supervised training scheme not only enables PatchNets to yield effective representations for local patches, but also allows for efficient PatchNet training with the help of global semantic labels. In addition, we construct a semantic codebook and propose a new encoding scheme, called as {\em vector of semantically aggregated descriptors} (VSAD), by exploiting the prediction score of PatchNet as posterior probability over semantic codewords. This VSAD encoding scheme overcomes the difficulty of dictionary learning in conventional methods like Fisher vector and VLAD, and produce more semantic and discriminative global representations. Moreover, we design a simple yet effective algorithm to select a subset of discriminative and representative codewords. This subset of codewords allows us to further compress the VSAD representation and reduce the computational cost on the large-scale dataset.

To verify the effectiveness of our proposed representations (i.e., PatchNet and VSAD), we focus on the problem of scene recognition. Specifically, we learn two PatchNets on two large-scale datasets, namely ImageNet~\cite{DengDSLL009} and Places~\cite{ZhouLXTO14}, and the resulted PacthNets denoted as {\bf object-PatchNet} and {\bf scene-PatchNet}, respectively. Due to the different training datasets, object-PatchNet and scene-PatchNet exhibit different but complementary properties, and allows us to develop more effective visual representations for scene recognition. As scene can be viewed as a collection of objects arranged in a certain spatial layout, we exploit the semantic probability of object-PatchNet to aggregate the features of the global pooling layer of scene-PatchNet. We conduct experiments on two standard scene recognition benchmarks (MIT Indoor67~\cite{QuattoniT09} and SUN397~\cite{XiaoHEOT10}) and the results demonstrate the superior performance of our VSAD representation to the current state-of-the-art approaches. Moreover, we comprehensively study different aspects of PatchNets and VSAD representations, aiming to provide more insights about our proposed new image representations for scene recognition.

The main contributions of this paper are summarized as follows:
\begin{itemize}
\item We propose a patch-level CNN to model the appearance of local patches, called as PatchNet. PatchNet is trained in a weakly-supervised manner simply with image-level supervision. Experimental results imply that PatchNet is more effective than classical image-level CNNs to extract semantic and discriminative features from local patches.
\item We present a new image representation, called as VSAD, which aggregates the PatchNet features from local patches with semantic probabilities. VSAD differs from previous CNN+FV for image representation on how to extract local features and how to estimate posterior probabilities for features aggregation. 
\item We exploit VSAD representation for scene recognition and investigate its complementarity to global CNN representations and traditional feature encoding methods. Our method achieves the state-of-the-art performance on the two challenging scene recognition benchmarks, i.e., MIT Indoor67 (86.2\%) and SUN397 (73.0\%), which outperforms previous methods with a large margin. The code of our method and learned models are made available to facilitate the future research on scene recognition. \footnote{\url{https://github.com/wangzheallen/vsad}}
\end{itemize}

The remainder of this paper is organized as follows. In Section~\ref{sec:rw}, we review related work to our method. After this, we briefly describe the Fisher vector representation to well motivate our method in Section~\ref{sec:vlad}. We present the PatchNet architecture and VSAD representation in Section~\ref{sec:method} and propose a codebook selection method in Section~\ref{sec:selection}. Then, we present our experimental results, verify the effectiveness of PatchNet and VSAD, and give a detailed analysis of our method in Section~\ref{sec:exp}. Finally, Section~\ref{sec:con} concludes this work.

\section{Related Work}
\label{sec:rw}

In this section we review related methods to our approach from the aspects of visual representation and scene recognition.

{\bf Visual representation.} 
Image recognition has received extensive research attention in past decades~\cite{KrizhevskySH12,HeZRS16,ZhouLXTO14,ShenLH15,SanchezPMV13,YangYGH09,ZhouYZH10,7529190,YuRT14,XuTHZ17,XuHZT16,LiuTSM17,liu2016large}. Early works focused on Bag of Visual Word representation~\cite{csurka2004,SivicZ03}, where local features were quantinized into a single word and a global histogram was utilized to summarize the visual content. Soft assigned encoding~\cite{GemertGVS08} method was introduced to reduce the information loss during quantization. Sparse coding~\cite{YangYGH09} and Locality-constrained linear coding~\cite{WangYYLHG10} was proposed to exploit sparsity and locality for dictionary learning and feature encoding. High dimensional encoding methods, such as Fisher vector~\cite{SanchezPMV13}, VLAD~\cite{JegouPDSPS12}, and Super Vector~\cite{ZhouYZH10}, was presented to reserve high-order information for better recognition. Our VSAD representation is mainly inspired by the encoding method of Fisher vector and VLAD, but differs in aspects of codebook construction and aggregation scheme.

Dictionary learning is another important component in image representation and feature encoding methods. Traditional dictionary (codebook) is mainly based on unsupervised learning algorithms, including $k$-means~\cite{csurka2004,SivicZ03}, Gaussian Mixture Models~\cite{SanchezPMV13}, $k$-SVD~\cite{1710377}. Recently, to enhance the discriminative power of dictionary,  several algorithms were designed for supervised dictionary learning~\cite{BoureauBLP10,SydorovSL14,PengWQP14}. Boureau {\em et al.}~\cite{BoureauBLP10} proposed a supervised dictionary learning method for sparse coding in image classification. Peng {\em et al.}~\cite{PengWQP14} designed a end-to-end learning to jointly optimize the dictionary and classifier weights for the encoding method VLAD. Sydorov {\em et al.}~\cite{SydorovSL14} presented a deep kernel framework and learn the parameters of GMM in a supervised way. The supervised GMMs were exploited for Fisher vector encoding. Wang {\em et al.}~\cite{WangWWQ16} proposed a set of good practices to enhance the codebook of VLAD representation. Unlike these dictionary learning method, the learning of our semantic codebook is weakly supervised with image-level labels transferred from the ImageNet dataset. We explicitly exploit object semantics in the codebook construction within our PatchNet framework. 

Recently Convolutional Neural Networks (CNNs)~\cite{lecun-98} have enjoyed great success for image recognition and many effective network architectures have been developed since the ILSVRC 2012~\cite{RussakovskyDSKS15}, such as AlexNet~\cite{KrizhevskySH12}, GoogLeNet~\cite{SzegedyLJSRAEVR15}, VGGNet~\cite{SimonyanZ14a}, and ResNet~\cite{HeZRS16}. These powerful CNN architectures have turned out to be effective for capturing visual representations for large-scale image recognition. In addition, several new optimization algorithms have been also proposed to make the training of deep CNNs easier, such as Batch Normalization~\cite{IoffeS15}, and Relay Back Propagation~\cite{ShenLH15}. Meanwhile, some deep learning architectures have been specifically designed for scene recognition~\cite{WangGHXQ16}. Wang {\em et al.}~\cite{WangGHXQ16} proposed a multi-resolution CNN architecture to capture different levels of information for scene understanding and introduced a soft target to disambiguate similar scene categories. Our PatchNet is a customized patch-level CNN to model local patches, while those previous CNNs aim to capture the image-level information for recognition.

There are several works trying to combine the encoding methods and deeply-learned representations for image and video recognition~\cite{arXiv:1601.07576,arXiv:1601.07977,DixitCGRV15,GongWGL14,ArandjelovicGTP15,Wang0T15}. These works usually were composed of two steps, where CNNs were utilized to extract descriptors from local patches and these descriptors were aggregated by traditional encoding methods. For instance, Gong {\em et al.} \cite{GongWGL14} employed VLAD to encode the activation features of fully-connected layers for image recognition. Dixit {\em et al.}~\cite{DixitCGRV15} designed a semantic Fisher vector to aggregate features from multiple layers (both convolutional and fully-connected layers) of CNNs for scene recognition. Guo {\em et al.}~\cite{arXiv:1601.07576} developed a locally-supervised training method to optimize CNN weights and proposed a hybrid representation for scene recognition. Arandjelovic {\em et al.}~\cite{ArandjelovicGTP15} developed a new generalized VLAD layer to train an end-to-end network for instance-level recognition. Our work is along the same research line of combining conventional and deep image representations. However, our method differs from these works on two important aspects: (1) we design a new PatchNet architecture to learn patch-level descriptors in a weakly supervised manner. (2) we develop a new aggregating scheme to summarize local patches (VSAD), which overcomes the limitation of unsupervised dictionary learning, and makes the final representation more effective for scene recognition.

{\bf Scene recognition.} 
Scene recognition is an important task in computer vision research~\cite{OlivaT01,WuR11,SongT10,YuTRC13,ObjectBank,ZhouLXTO14,WangGHXQ16} and has many applications such as event recognition~\cite{XiongZLT15,WangWDQ15} and action recognition~\cite{Wang0T16,Wang0TG16,WangQT14}. Early methods made use of hand-crafted global features, such as GIST~\cite{OlivaT01}, for scene representation. Global features are usually extracted efficiently to capture the holistic structure and content of the entire image. Meanwhile, several local descriptors (e.g., SIFT \cite{Lowe04}, HOG~\cite{DalalT05}, and CENTRIST~\cite{WuR11}) have been developed for scene recognition within the frameworks of Bag of Visual Words (e.g., Histogram Encoding~\cite{SivicZ03}, Fisher vector \cite{PerronninSM10}). These representations leveraged information of local regions for scene recognition and obtained good performance in practice. However, local descriptors only exhibit limited semantics and so several mid-level and high-level representations have been introduced to capture the discriminative parts of scene content (e.g., mid-level patches~\cite{SinghGE12}, distinctive parts~\cite{JunejaVJZ13}, object bank~\cite{ObjectBank}). These mid-level and high-level representations were usually discovered in an iterative way and trained with a discriminative SVM. Recently, several structural models were proposed to capture the spatial layout among local features, scene parts, and containing objects, including spatial pyramid matching~\cite{SPM}, deformable part based model~\cite{DPMscene}, reconfigurable models~\cite{PariziOF12}. These structural models aimed to describe the structural relation among visual components for scene understanding.

\begin{table*}
\centering
\small
\caption{\textbf{PatchNet Architecture:} We adapt the successful Inception V2~\cite{IoffeS15} structure to the design of PatchNet, which takes a $128 \times 128$ image region as input and outputs its semantic probability. In experiment, we also study the performance of PatchNet with VGGNet16~\cite{SimonyanZ14a} structure.}
\resizebox{\textwidth}{!}{
\begin{tabular}{|c|c|c|c|c|c|c|c|}
  \hline
  Layer & conv1  & conv2 & Inception3a & Inception3b & Inception3c & Inception4a & Inception4b  \\
  \hline
  Feature map size & $64 \times 64 $ & $32 \times 32 $& $16 \times 16$ & $16 \times 16 $ & $8\times8$  & $8\times8$  &$8\times8$   \\
  \hline
  Stride & 2 & 1 & 1 & 1 & 2 & 1 & 1   \\
  \hline
  Channel & 64 & 192 & 256 & 320 & 576 & 576 & 576  \\
  \hline
  Layer map & Inception4c & Inception4d & Inception4e & Inception5a & Inception5b & global Avg & prediction  \\
  \hline
  Feature map size & $8\times8$  & $8 \times 8$ & $4 \times 4$ & $4 \times 4$ &  $4 \times 4$ & $1 \times 1$ & $1 \times 1$  \\
  \hline
  Stride & 1 & 1 & 2 & 1 & 1 & 1 & 1  \\
  \hline
  Channel & 608 & 608 & 1056 & 1024 & 1024 & 1024 & 1000  \\
  \hline
\end{tabular}
}
\label{tbl:convnet}
\end{table*}

Our PatchNet and VSAD representations is along the research line of exploring more semantic parts and objects for scene recognition. Our method has several important differences from previous scene recognition works: (1) we utilize the recent deep learning techniques (PatchNet) to describe local patches for CNN features and aggregate these patches according to their semantic probabilities. (2) we also explore the general object and scene relation to discover a subset of object categories to improve the representation capacity and computational efficiency of our VSAD.

\section{Motivating PatchNets}
\label{sec:vlad}
In this section, we first briefly revisit Fisher vector method. Then, we analyze the Fisher vector representation to well motivate our approach. 

\subsection{Fisher Vector Revisited}

Fisher vector~\cite{SanchezPMV13} is a powerful encoding method derived from Fisher kernel and has proved to be effective in various tasks such as object recognition~\cite{PerronninSM10}, scene recognition~\cite{JunejaVJZ13}, and action recognition~\cite{WangWQ12,PengWWQ16}. Like other conventional image representations, Fisher vector aggregates local descriptors into a global high-dimensional representation. Specifically, a Gaussian Mixture Model (GMM) is first learned to describe the distribution of local descriptors. Then, the GMM posterior probabilities are utilized to softly assign each descriptor to different mixture components. After this, the first and second order differences between local descriptors and component center are aggregated in a weighted manner over the whole image. Finally, these difference vectors are concatenated together to yield the high-dimensional Fisher vector ($2KD$), where $K$ is the number of mixture components and $D$ is the descriptor dimension.

\subsection{Analysis}

From the above description about Fisher vector, there are two key components in this aggregation-based representation:
\begin{itemize}
	\item The first key element in Fisher vector encoding method is the local descriptor representation, which determines the feature space to learn GMMs and aggregate local patches.
	\item The generative GMM is the second key element, as it defines a soft partition over the feature space and determines how to aggregate local descriptors according to this partition.
\end{itemize}

Conventional image representations rely on hand-crafted features, which may not be optimal for classification tasks, while recent methods~\cite{GongWGL14,DixitCGRV15} choose image-level deep features to represent local patches, which are not designed for patch description by its nature. Additionally, dictionary learning (GMM) method heavily relies on the design of patch descriptor and its performance is highly correlated with the choice of descriptor. Meanwhile, dictionary learning is often based on unsupervised learning algorithms and sensitive to the initialization. Moreover, the learned codebook lacks semantic property and it is hard to interpret and visualize these mid-level codewords. These important issues motivate us to focus on two aspects to design effective visual representations: (1) {\em how to describe local patches with more powerful and robust descriptors;} and (2) {\em how to aggregate these local descriptors with more semantic codebooks and effective schemes}.

\section{Weakly Supervised PatchNets}
\label{sec:method}

\begin{table}
\caption{Summary of notations used in our method.}
\begin{tabular}{l|l}
\hline
$\mathbf{x}$ & local patch sampled from images \\
$\mathbf{z}$ & latent variable to model local patches \\
$\mathbf{f}$ & patch-level descriptor extracted with PatchNet \\
$\mathbf{p}$ & patch-level semantic probability extracted PatchNet \\
$\mathcal{X}$ & a set of local patches \\
$\mathcal{F}$ & a set of patch descriptors \\
$\mathcal{P}$ & a set of semantic probability distributions \\
$\mathbf{p}_{image}$ & image-level semantic probability \\
$\mathbf{p}_{category}$ & semantic probability over images from a category \\
$\mathbf{p}_{data}$ & semantic probability over all images from a dataset \\
\hline
\end{tabular}
\end{table}

In this section we describe the PatchNet architecture to model the appearance of local patches and aggregate them into global representations. First, we introduce the network structure of PatchNet. Then, we describe how to use learned PatchNet models to describe local patches. Finally, we develop a semantic encoding method (VSAD) to aggregate these local patches, yielding the image-level representation.

\subsection{PatchNet Architectures}

The success of aggregation-based encoding methods (e.g., Fisher vector~\cite{PerronninSM10}) indicates that the patch descriptor is a kind of rich representation for image recognition. A natural question arises that whether we are able to model the appearance of these local patches with a deep architecture, that is trainable in an end-to-end manner. However, the current large-scale datasets (e.g., ImageNet~\cite{DengDSLL009}, Places~\cite{ZhouLXTO14}) simply provide the image-level labels without the detailed annotations of local patches. Annotating every patch is time-consuming and sometimes could be ambiguous as some patches may contain part of objects or parts from multiple objects. To handle these issues, we propose a new patch-level architecture to model local patches, which is still trainable with the image-level labels.

Concretely, we aim to learn the patch-level descriptor directly from raw RGB values, by classifying them into predefined semantic categories (e.g., object classes, scene classes). In practice, we apply the image-level label to each randomly selected patch from this image, and utilize this transferred label as supervision signal to train the PatchNet. In this training setting, we do not have the detailed patch-level annotations and exploit the image-level supervision signal to learn patch-level classifier. So, the PatchNet could be viewed as a kind of weakly supervised network. We find that although the image-level supervision may be inaccurate for some local patches and the converged training loss of PatchNet is higher than that of image-level CNN, it is still able to learn effective representation to describe local patches and reasonable semantic probability to aggregate these local patches.

Specifically, our proposed PatchNet is a CNN architecture taking small patches ($ 128 \times 128 $) as inputs. We adapt two famous image-level structures (i.e., VGGNet~\cite{SimonyanZ14a} and Inception V2~\cite{IoffeS15}) for the PatchNet design. The Inception based architecture is illustrated in Table \ref{tbl:convnet}, and its design is inspired by the successful Inception V2 model with batch normalization~\cite{IoffeS15}. The network starts with 2 convolutional and max pooling layers, subsequently has 10 inception layers, and ends with a global average pooling layer and fully connected layer. Different from the original Inception V2 architecture, our final global average pooling layer has a size of $ 4 \times 4 $ due to the smaller input size ($ 128 \times 128 $). The output of PatchNet is to predict the semantic labels specified by different datasets (e.g., 1,000 object classes on the ImageNet dataset, 205 scene classes on the Places dataset). In practice, we train two kinds of PatchNets: {\bf object-PatchNet} and {\bf scene-PatchNet}, and the training details will be explained in subsection~\ref{sec:detail}.

{\bf Discussion.} Our PatchNet is a customized network for patch modeling, which differs from the traditional CNN architectures on two important aspects: (1) our network is a patch-level structure and its input is a smaller image region ($128 \times 128$) rather than a image ($ 224 \times 224 $), compared with those image-level CNNs~\cite{KrizhevskySH12,SimonyanZ14a,SzegedyLJSRAEVR15}; (2) our network is trained in a weakly supervised manner, where we directly treat the image-level labels as patch-level supervision information. Although this strategy is not accurate, we empirically demonstrate that it still enables our PacthNet to learn more effective representations for aggregation-based encoding methods in our experiments. 

\begin{figure*}[t]
\includegraphics[width=\textwidth]{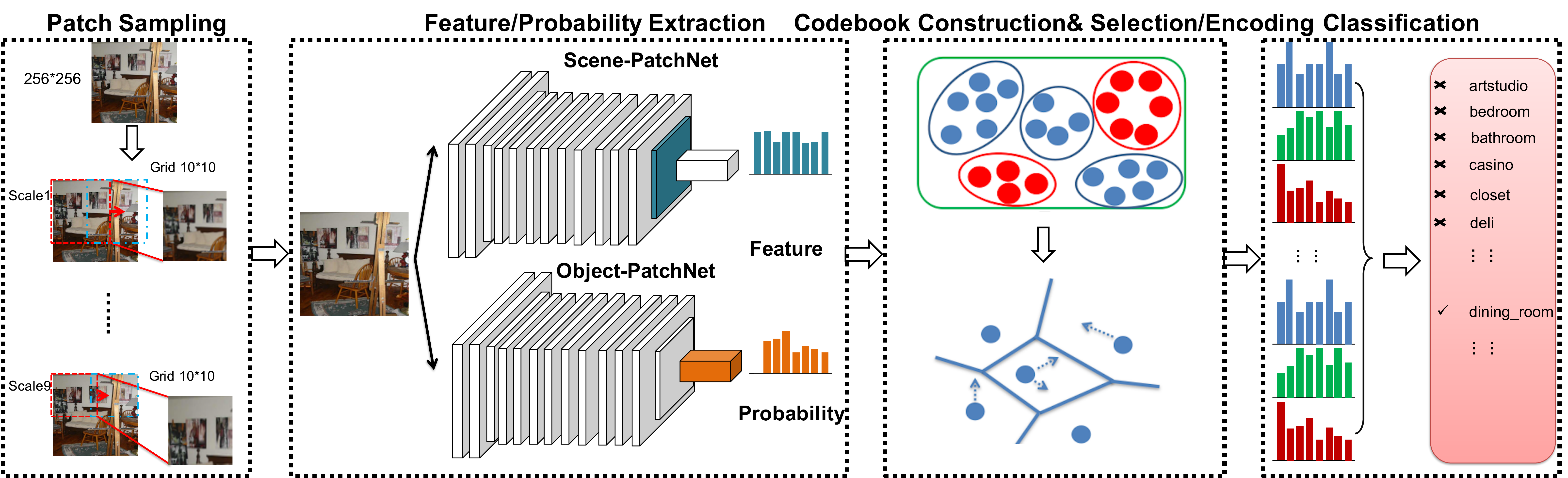}
\caption{{\bf Pipeline of our method}. We first densely sample local patches in a multi-scale manner. Then, we utilize two kinds of PatchNets to describe each patch (Scene-PatchNet feature) and aggregate these patches (Object-PatchNet probability). Based on our learned semantic codebook, these local patches are aggregated into a global representation with VSAD encoding scheme. Finally, these global representations are exploited for scene recognition with a linear SVM.}
\end{figure*}

\subsection{Describing Patches}

After the introduction of PatchNet architecture, we are ready to present how to describe local patches with PatchNet. The proposed PatchNet is essentially a patch-level discriminative model, which aims to map these local patches from raw RGB space to a semantic space determined by the supervision information. PatchNet is composed of a set of standard convolutional and pooling layers, that process features with more abstraction and downsample spatial dimension to a lower resolution, capturing full content of local patches. During this procedure, PatchNet hierarchically extracts multiple-level representations (hidden layers, denoted as $\mathbf{f}$) from raw RGB values of patches, and eventually outputs the probability distribution over semantic categories (output layers, denoted as $\mathbf{p}$). 

The final semantic probability $\mathbf{p}$ is the most abstract and semantic representation of a local patch. Compared with the semantic probability, the hidden layer activation features $\mathbf{f}$ are capable of containing more detailed and structural information. Therefore, multiple-level representations $\mathbf{f}$ and semantic probability $\mathbf{p}$ could be exploited in two different manners: {\em describing} and {\em aggregating} local patches. In our experiments, we use the activation features of the last hidden layer as the patch-level descriptors. Furthermore, in practice, we could even try the combination of activation features $\mathbf{f}$ and semantic probability $\mathbf{p}$ from different PatchNets (e.g., object-PatchNet, scene-PatchNet). This flexible scheme decouples the correlation between local descriptor design and dictionary learning, and allows us to make best use of different PatchNets for different purposes according to their own properties. 

\subsection{Aggregating Patches}

After having introduced the architecture of PatchNet to describe the patches with multiple-level representations $\mathbf{f}$ in the previous subsection, we present how to aggregate these patches with semantic probability $\mathbf{p}$ of PatchNet in this subsection. As analyzed in Section~\ref{sec:vlad}, aggregation-based encoding methods (e.g., Fisher vector) often rely on generative models (e.g., GMMs) to calculate the posterior distribution of a local patch, indicating the probability of belonging to a codeword. In general, the generative model often introduces latent variables $\mathbf{z}$ to capture the underline factors and the complex distribution of local patches $\mathbf{x}$ can be obtained by marginalization over latent variables $\mathbf{z}$ as follows:
\begin{equation}
	p(\mathbf{x}) = \sum_{\mathbf{z}} p(\mathbf{x} | \mathbf{z}) p(\mathbf{z}).
\end{equation}

However, from the view of aggregation process, only the posterior probability $p(\mathbf{z} | \mathbf{x})$ are needed to assign a local patch $\mathbf{x}$ to these learned codewords in a soft manner. Thus, it will not be necessary to use generative model $p(\mathbf{x})$ for estimating $p(\mathbf{z} | \mathbf{x})$, and we can directly calculate $p(\mathbf{z} | \mathbf{x})$ with our proposed PatchNet.  Directly modeling posterior probability with PatchNet exhibits two advantages over traditional generative models:
\begin{itemize}
	\item The estimation of $p(\mathbf{x})$ is a non-trivial task and the learning of generative models (e.g., GMMs) is sensitive to the initialization and may converge to a local minimum. Directly modeling $p(\mathbf{z} | \mathbf{x})$ with PatchNets can avoid this difficulty by training on large-scale supervised datasets.
    \item Prediction scores of PatchNet correspond to semantic categories, which is more informative and semantic than that of the original generative model (e.g., GMMs). Utilizing this semantic posterior probability enables the final representation to be interpretable.
\end{itemize}

\begin{figure*}[t]
	\includegraphics[width=\textwidth]{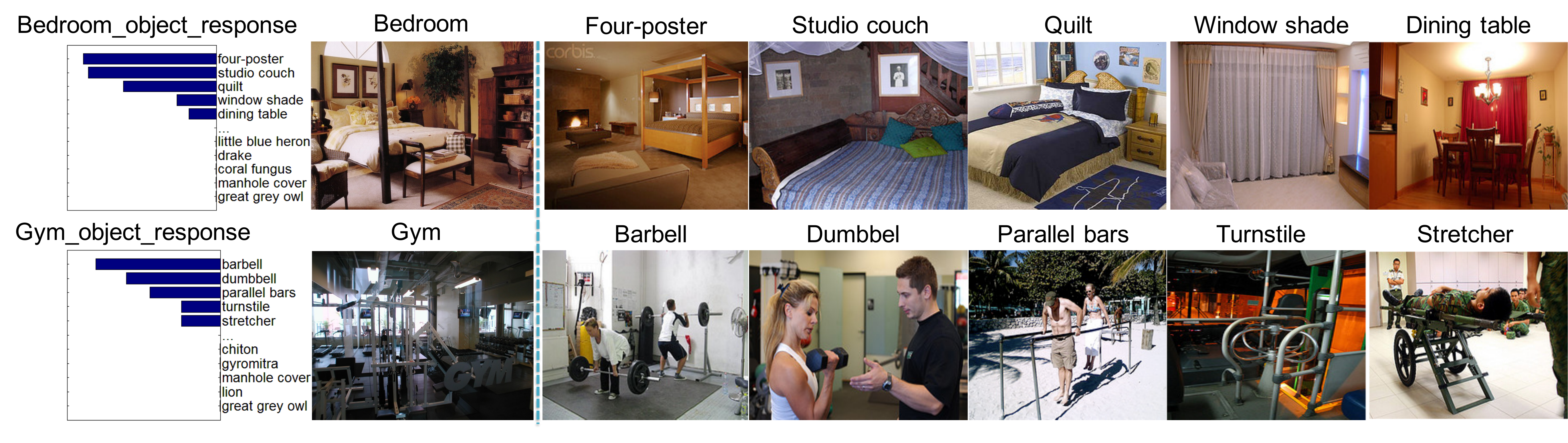}
	\caption{{\bf Illustration of scene-object relationship}. The first row is the \emph{Bedroom} scene with its top 5 most likely object classes. Specifically, we feed all the training image patches of the \emph{Bedroom} scene into our PatchNet.
For each object category, we sum over the conditional probability over all training patches as the response for this object.
The results are shown in the 1st column. We then show five object classes (top 5 objects) for the \emph{Bedroom} scene (the second to the sixth column). The second row is an illustration for the \emph{Gym} scene, which is a similar case to \emph{Bedroom}.}
    \label{fig:scene_object_response}
\end{figure*}

{\bf Semantic codebook.} We first describe the semantic codebook construction based on the semantic probability extracted with PatchNet. In particular, given a set of local patches $\mathcal{X} = \{\mathbf{x}_1, \mathbf{x}_2, \ldots, \mathbf{x}_N\}$, we first compute their semantic probabilities with PatchNet, denoted as $\mathcal{P} = \{\mathbf{p}_1, \mathbf{p}_2, \ldots, \mathbf{p}_N\}$. We also use PatchNet to extract patch-level descriptors $\mathcal{F} = \{\mathbf{f}_1, \mathbf{f}_2, \ldots, \mathbf{f}_N\}$. Finally, we generate semantic mean (center) for each codeword as follows:
\begin{equation}
	\mu_k = \frac{1}{N_k}\sum_{i=1}^N \mathbf{p}_i^k \mathbf{f}_i,
\label{equ:mean}
\end{equation}
where $\mathbf{p}_i^k$ is the $k^{th}$ dimension of $\mathbf{p}_i$, and $N_k$ is calculated as follows:
\begin{equation}
	N_k = \sum_{i=1}^N \mathbf{p}_i^k, \ \ \ \pi_k = \frac{N_k}{N}.
\label{equ:pi}
\end{equation}
We can interpret $N_k$ as the prior distribution over the semantic categories and $\mu_k$ as the category template in this feature space $\mathbf{f}$. Meanwhile, we can calculate the semantic covariance for each codeword by the following formula:
\begin{equation}
	\Sigma_k = \frac{1}{N_k}\sum_{i=1}^N \mathbf{p}_i^k (\mathbf{f}_i - \mu_k) (\mathbf{f}_i - \mu_k)^\top.
\label{equ:covariance}
\end{equation}
The semantic mean and covariance in Equation (\ref{equ:mean}) and (\ref{equ:covariance}) constitute our semantic codebook, and will be exploited to semantically aggregate local descriptors in the next paragraph.

{\bf VSAD.} After the description of PatchNet and semantic codebook, we are able to develop our hybrid visual representations, namely {\em vector of semantically aggregating descriptor} (VSAD). Similar to Fisher vector \cite{PerronninSM10}, given a set of local patches with descriptors $\{\mathbf{f}_1, \mathbf{f}_2, \ldots, \mathbf{f}_T\}$, we aggregate both first order and second order information of local patches with respect to semantic codebook as follows:
\begin{equation}
\mathcal{S}_{k} = \frac{1}{{\sqrt {{\pi_k}} }}\sum\limits_{t = 1}^T {{\mathbf{p}_t^k}} \left( {\frac{{{\mathbf{f}_t} - {\mu_k}}}{{{\sigma_k}}}} \right),
\label{equ:mean_vsad}
\end{equation}
\begin{equation}
\mathcal{G}_{k} = \frac{1}{{\sqrt {{\pi_k}} }}\sum\limits_{t = 1}^T {{\mathbf{p}_t^k}} \left[ {\frac{{{{({\mathbf{f}_t} - {\mu_k})}^2}}}{{\sigma_k^2}} - 1} \right],
\label{equ:var_vsad}
\end{equation}
where $\{\pi, \mu, \sigma\}$ is semantic codebook defined above, $\mathbf{p}$ is the semantic probability calculated from PatchNet, $\mathcal{S}$ and $\mathcal{G}$ are first and second order VSAD, respectively. Finally, we concatenate these sub-vectors from different codewords to form our VSAD representation: $[\mathcal{S}_1, \mathcal{G}_1, \mathcal{S}_2, \mathcal{G}_2, \cdots, \mathcal{S}_K, \mathcal{G}_K]$.

\section{Codeword Selection for Scene Recognition}
\label{sec:selection}

In section we take scene recognition as a specific task for image recognition and utilize object-PatchNet for semantic codebook construction and VSAD extraction. Based on this setting, we propose an effective method to discover a set of discriminative object classes to compress VSAD representation. It should be noted that our selection method is general and could be applied to other relevant tasks and PatchNets.

\begin{figure*}[t]
	\includegraphics[width=\textwidth]{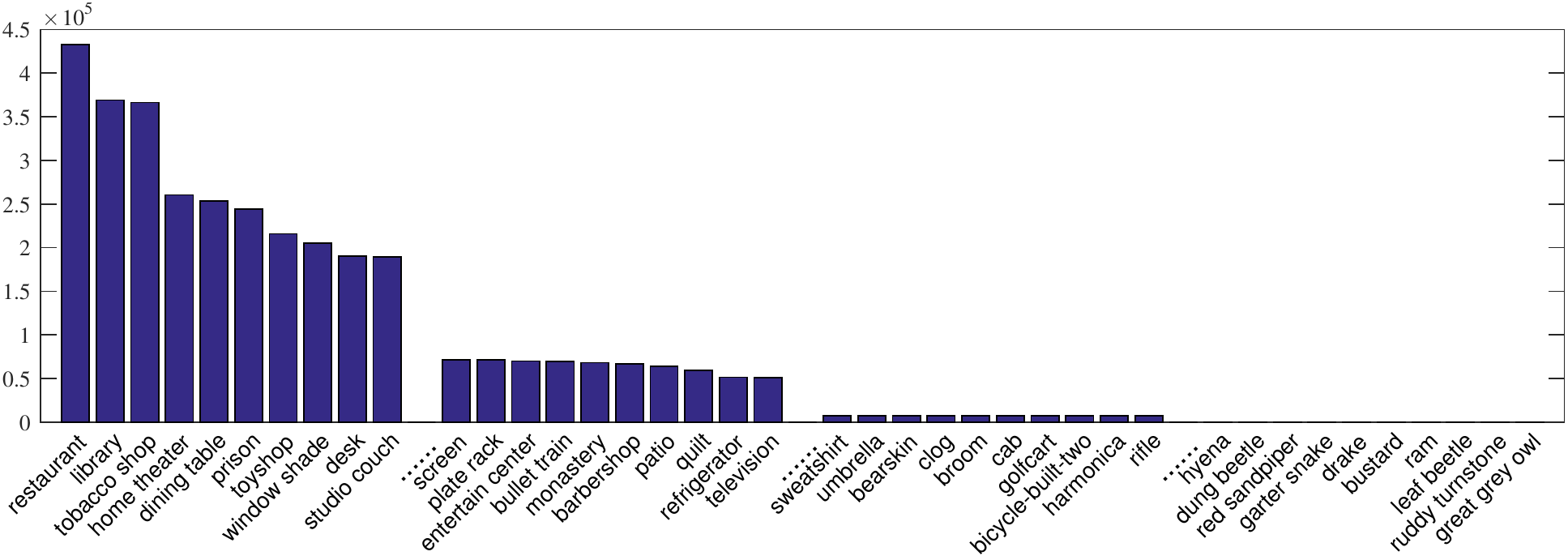}
\caption{ {\bf Illustration of the object responses in the object-PatchNet}. Specifically, we feed all the training patches (MIT Indoor 67) into our object-PatchNet, and obtain the corresponding probability distribution for each patch. For each object category, we use the sum of probabilities over all the training patches as the response of this object category. Then we sort the responses of all the object categories in a descent order. For visual clarity, we here show four typical groups with high (from \emph{restaurant} to \emph{studio couch}), moderate (from \emph{screen} to \emph{television}),
minor (from \emph{sweatshirt} to \emph{rifle}), and low responses (from \emph{hyena} to \emph{great grey owl}). We can see that the groups with the minor and low response (the response rank of these objects: around 250 to 1000) make very limited contribution to the whole scene dataset. Hence, we should design our selection strategy to discard them to reduce the redundance of our semantic codebook.}    
\label{fig:overall_patchdescription}
\end{figure*}

Since our semantic codebook is constructed based on the semantic probability $\mathbf{p}$ of object-PatchNet, the size of our codebook is equal to the number of object categories from our PatchNet (i.e., 1000 objects in ImageNet). However, this fact may reduce the effectiveness of our VSAD representation due to the following reasons:
\begin{itemize}
	\item Only a few object categories in ImageNet are closely related with scene category. In this case, many object categories in our semantic codebook are redundant. We here use the \emph{Bedroom} and 		\emph{Gym} scene classes (from MIT Indoor67~\cite{QuattoniT09}) as an illustration for scene-object relationship. As shown in Figure \ref{fig:scene_object_response}, we can see that the \emph{Bedroom} scene 		class most likely contains the object classes \emph{Four-poster}, \emph{Studio couch}, \emph{Quilt}, \emph{Window shade}, \emph{Dining table}. The \emph{Gym} scene class is a similar case. Furthermore, we 	feed all the training patches of MIT Indoor 67 into our object-PatchNet. For each object category, we sum over the conditional probability of all the training patches as the response for this object. The 		result in Figure \ref{fig:overall_patchdescription} indicates that around 750 categories of 1000 are not activated. Hence, the redundance using 1,000 object categories is actually large.
	\item From the computational perspective, the large size of codebook will prohibit the application of VSAD on large-scale datasets due to the huge consumption of storage and memory. Therefore, it is also necessary to select a subset of codewords (object categories) to compress the VSAD representation and improve the computing efficiency.
\end{itemize}

Hence, we propose a codeword selection strategy as follows to enhance the efficiency of our semantic codebook and improve the computation efficiency of our VSAD representation.
Specifically,
we take advantage of the scene-object relationship to select $K$ classes of 1000 ImageNet objects for our semantic codebook generation.
\textbf{First},
the probability vector $\mathbf{p}_{patch}$ of the object classes for each training patch is obtained from the output of our PatchNet.
We then compute the response of the object classes
for each training image $\mathbf{p}_{image}$, each scene category $\mathbf{p}_{category}$ and the whole training data $\mathbf{p}_{data}$
\begin{align}
\mathbf{p}_{image} &= \sum\nolimits_{patch \in image} \mathbf{p}_ {patch}\\
\mathbf{p}_{category} &= \sum\nolimits_{image \in category} \mathbf{p}_ {image}\\
\mathbf{p}_{data} &= \sum\nolimits_{catrgory \in data} \mathbf{p}_ {category}
\end{align}
\textbf{Second},
we rank $p_{data}$ in the descending order and select $2K$ object classes (with top $2K$ highest responses).
We denote the resulting object set as $\mathcal{O}_{data} = \{o_j\}_{j=1}^{2K}$.
\textbf{Third},
for each scene category,
we rank $p_{category}$ in the descending order and select $T$ object classes (with top $T$ highest responses).
Then we collect the object classes for all the scene categories together, and delete the duplicate object classes.
We denote the object set as $\mathcal{O}_{category} = \{o_i\}_{i=1}^{M}$, where $M$ is the number of object classes in $\mathcal{O}_{category}$.
\textbf{Finally},
the intersection of $\mathcal{O}_{category}$ and $\mathcal{O}_{data}$ is used as the selected object class set, i.e., $\mathcal{O} \leftarrow \mathcal{O}_{category} \cap \mathcal{O}_{data}$.
To constrain the number of object classes as the predefined $K$, we can gradually increase $T$ (when selecting $\mathcal{O}_{category}$), starting from one.
Additionally, to speed up the selection procedure, we choose $2K$ as the size of $\mathcal{O}_{data}$.
Note that,
our selected object set $O$ is the intersection of $\mathcal{O}_{category}$ and $\mathcal{O}_{data}$.
In this case, the selected object classes not only contain the general characteristics of the entire scene dataset, but also the specific characteristics of each scene category.
Consequentially,
this selection strategy enhances the discriminative power of our semantic codebook and VSAD representations, yet is still able to reduce the computational cost.

\section{Experiments}
\label{sec:exp}

In this section we evaluate our method on two standard scene recognition benchmarks to demonstrate its effectiveness. First, we introduce the evaluation datasets and the implementation details of our method. Then, we perform exploration experiments to determine the important parameters of the VSAD representation. Afterwards, we comprehensively study the performance of our proposed PatchNets and VSAD representations. In addition, we also compare our method with other state-of-the-art approaches. Finally, we visualize the semantic codebook and the scene categories with the most performance improvement.

\subsection{Evaluation Datasets and Implementation Details}
\label{sec:detail}

\begin{figure*}[t]
\centering
\includegraphics[width=0.33\linewidth]{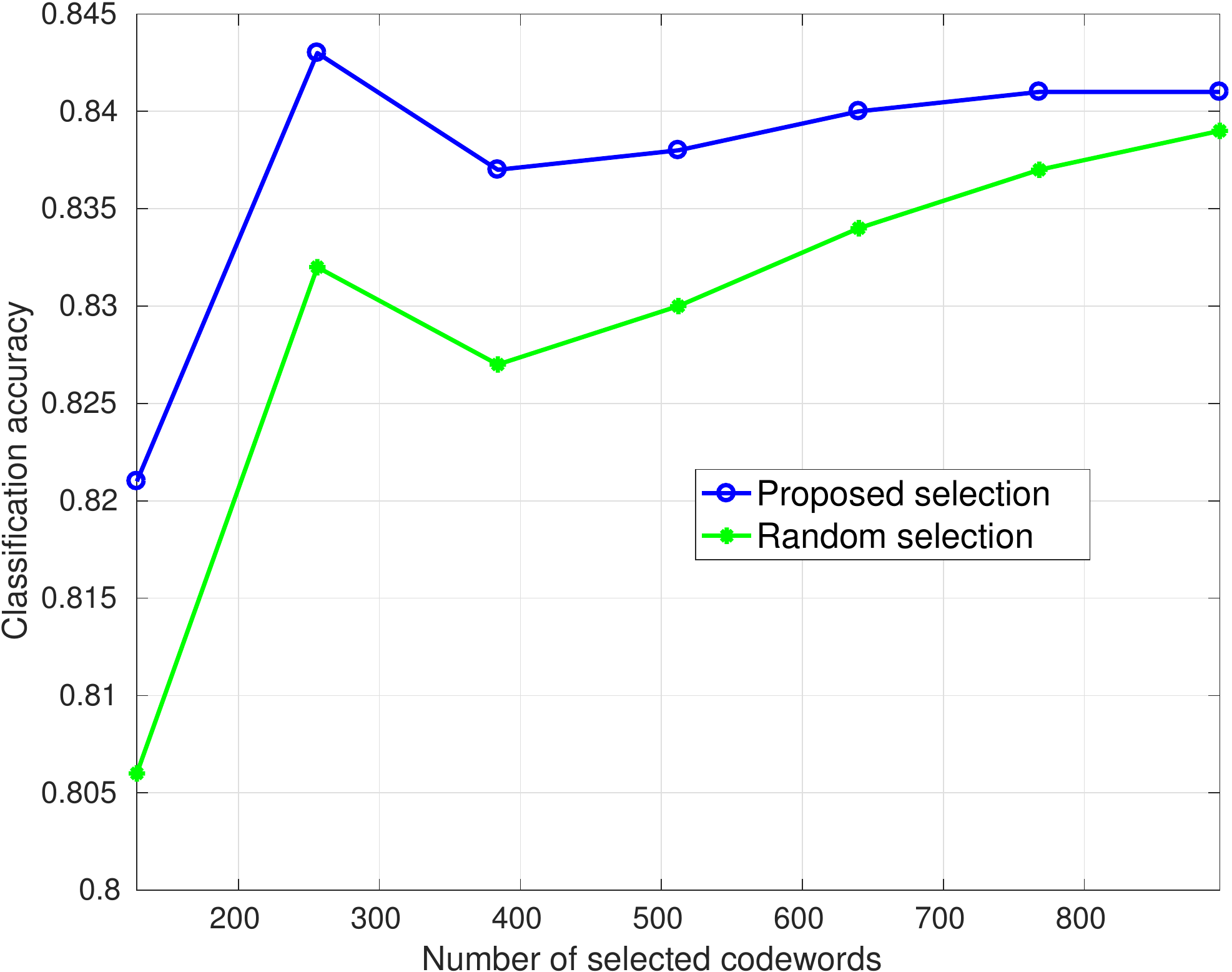}
\includegraphics[width=0.32\linewidth]{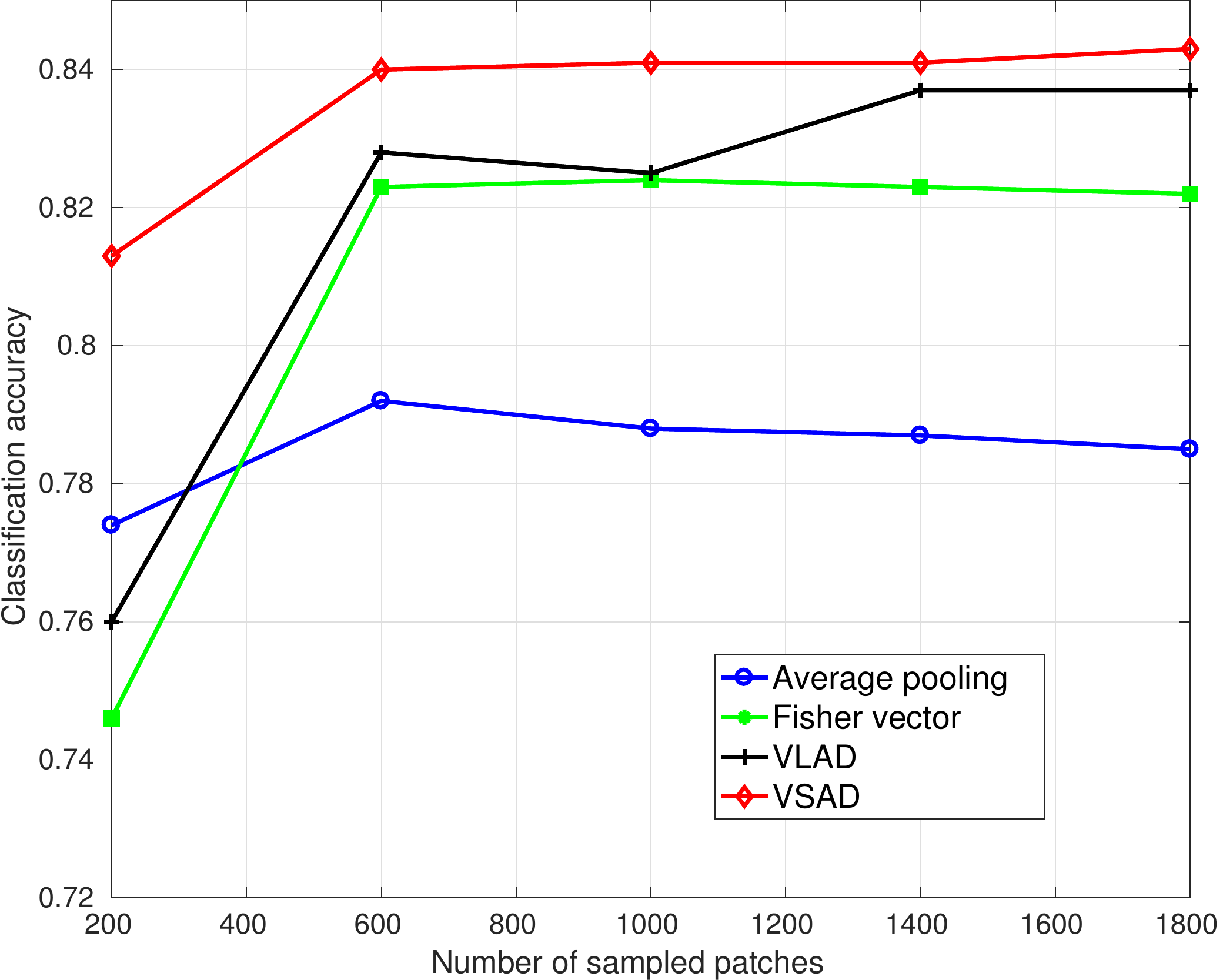}
\includegraphics[width=0.328\linewidth]{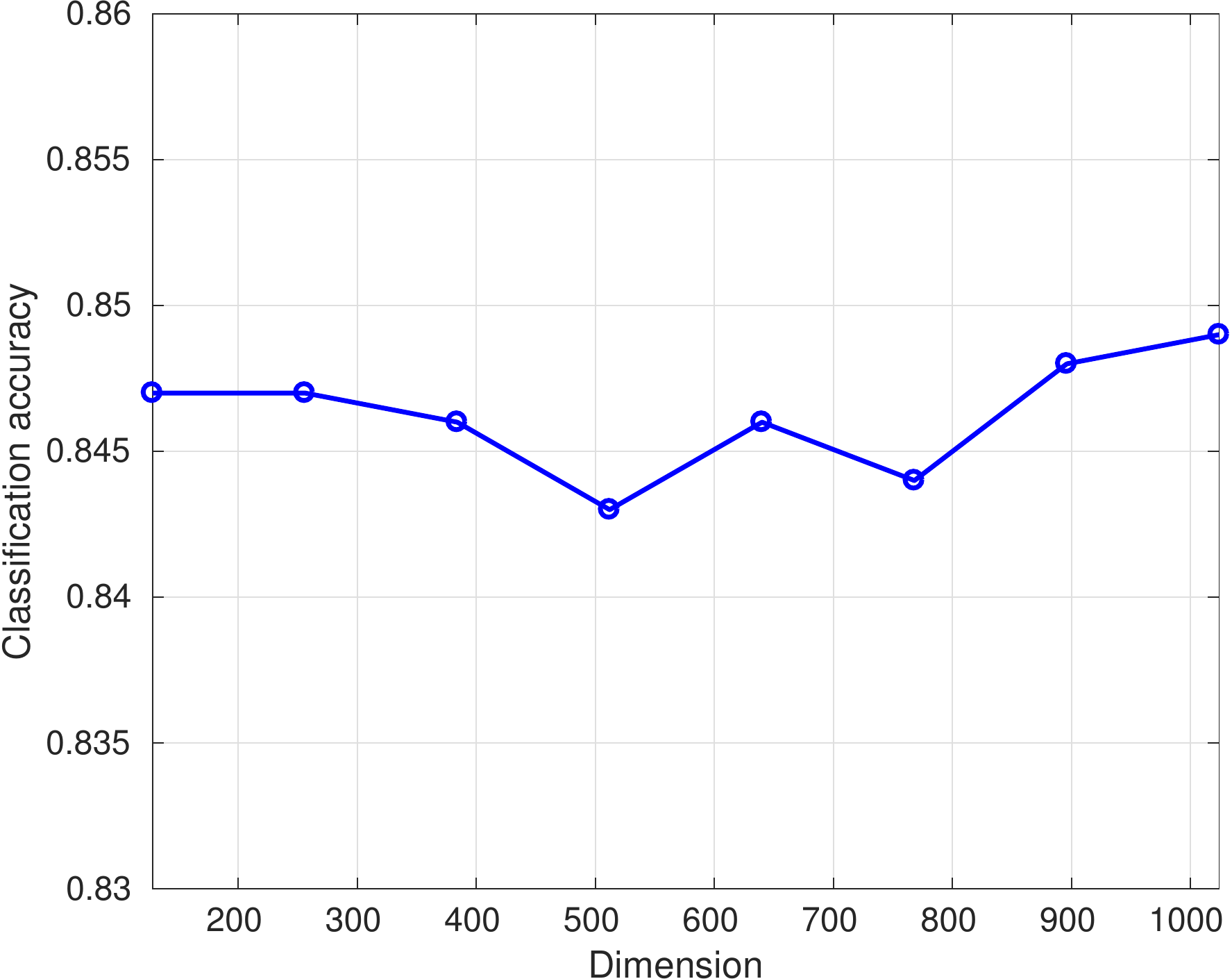}
\caption{{\bf Exploration study on the MIT Indoor67 dataset}. Left: performance comparison of different codebook selection methods; Middle: performance comparison of different numbers of sampled patches; Right: performance comparison of different descriptor dimension reduced by PCA.}
\label{fig:exploration}
\end{figure*}

Scene recognition is a challenging task in image recognition, due to the fact that scene images of the same class exhibit large intra-class variations, while images from different categories contain small inter-class differences. Here, we choose this challenging problem of scene recognition as the evaluation task to demonstrate the effectiveness of our proposed PatchNet architecture and VSAD representation. Additionally, scene image can be viewed as a collection of objects arranged in the certain layout, where the small patches may contain rich object information and can be effectively described by our PatchNet. Thus scene recognition is more suitable to evaluate the performance of VSAD representation.

\textbf{Evaluation datasets.} In our experiment, we choose two standard scene recognition benchmarks, namely MIT Indoor67~\cite{QuattoniT09} and SUN397~\cite{XiaoHEOT10}. The MIT Indoor67 dataset contains 67 indoor-scene classes and has 15,620 images in total. Each scene category contains at least 100 images, where 80 images are for training and 20 images for testing. The SUN397 dataset is a larger scene recognition dataset, including 397 scene categories and 108,754 images, where each category also has at least 100 images. We follow the standard evaluation from the original paper~\cite{XiaoHEOT10}, where each category has 50 images for training and 50 images for testing. Finally, the average classification accuracy over 10 splits is reported.

\textbf{Implementation details of PatchNet training.} In our experiment, to fully explore the modeling power of PatchNet, we train two types of PatchNets, namely {\bf scene-PatchNet} and {\bf object-PatchNet} with the MPI extension~\cite{WangXW0LTG16} of Caffe toolbox~\cite{jia2014caffe}. The scene-PatchNet is trained on the large-scale Places dataset~\cite{ZhouLXTO14}, and the object-PatchNet is learned from the large-scale ImageNet dataset~\cite{DengDSLL009}. The Places dataset contains around 2,500,000 images and 205 scene categories and the ImageNet dataset has around 1,300,000 images and 1,000 object categories. We train both PatchNets from scratch on these two-large scale datasets. Specifically, we use the stochastic gradient decent (SGD) algorithm to optimize the model parameters, where momentum is set as 0.9 and batch size is set as 256. The learning rate is initialized as $0.01$ and decreased to its $\frac{1}{10}$ every $K$ iterations. The whole learning process stops at $3.5K$ iterations. $K$ is set as 200,000 for the ImageNet dataset and 350,000 for the Places dataset. To reduce the effect of over-fitting, we adopt the common data augmentation techniques. We first resize each image into size of $256 \times 256$. Then we randomly crop a patch of size $s \times s$ from each image, where $s \in \{64, 80, 96, 112, 128, 144, 160, 176, 192\}$. Meanwhile, these cropped patches are horizontally flipped randomly. Finally, these cropped image regions are resized as $128 \times 128$ and fed into PatchNet for training. The object-PatchNet achieves the recognition performance of $\mathbf{85.3\%}$ (top-5 accuracy) on the ImageNet dataset and the scene-PatchNet obtains the performance of $\mathbf{82.7\%}$ (top-5 accuracy) on the Places dataset.

\textbf{Implementation details of patch sampling and classifier.} An important implementation detail in the VSAD representation is how to densely sample patches from the input image. To deal with the large intra-class variations existed in scene images, we design a {\em multi-scale dense sampling} strategy to select image patches. Specifically, like training procedure, we first resize each image to size of $256 \times 256$. Then, we sample patches of size $s \times s $ from the whole image in the grid of $10 \times 10$. Sizes $s$ of these sampled patches range from $\{64, 80, 96, 112, 128, 144, 160, 176, 192\}$. These sampled image patches also go under horizontal flipping for further data augmentation. Totally, we have 9 different scales and each scale we sample 200 patches ($10 \times 10$ grid and $2$ horizontal flips). Normalization and recognition classifier are other important factors for all encoding methods (i.e., average pooling, VLAD, Fisher vector, and VSAD). In our experiment, the image-level representation is signed-square-rooted and L2-normalized for all encoding methods. For classification, we use a linear SVM (C=1) trained in the one-vs-all setting. The final predicted class is determined by the maximum score of different binary SVM classifiers.

\subsection{Exploration Study}

In this subsection we conduct exploration experiments to determine the parameters of important components in our VSAD representation. First, we study the performance of our proposed codeword selection algorithm and determine how many codewords are required to construct efficient VSAD representation. Then, we study the effectiveness of proposed multi-scale sampling strategy and determine how many scales are needed for patch extraction. Afterwards, we conduct experiments to explore the dimension reduction of PatchNet descriptors. Finally, we study the influence of different network structures and compare Inception V2 with VGGNet16. In these exploration experiments, we choose scene-PatchNet to describe each patch (i.e., extracting descriptors $\mathbf{f}$), and object-PatchNet to aggregate patches (i.e., utilizing semantic probability $\mathbf{p}$). We perform this exploration experiment on the dataset of MIT Indoor67.

\textbf{Exploration on codeword selection.}
We begin our experiments with the exploration of codeword selection. We propose a selection strategy to choose the number of object categories (the codewords of semantic codebook) in Section \ref{sec:selection}. We report the performance of VSAD representation with different codebook sizes in the left of Figure \ref{fig:exploration}. To speed up this exploration experiment, we use PCA to pre-process the patch descriptor $\mathbf{f}$ by reducing its dimension from 1,024 to 100. In our study, we compare the performance of our selection method with the random selection. As expected, our selection method outperforms the random selection, in particular when the number of selected codewords are small. Additionally, when selecting 256 codewords, we can already achieve a relatively high performance. Therefore, to keep a balance between recognition performance and computing efficiency, we fix the number of selected codewords as 256 in the remaining experiments.

\textbf{Exploration on multi-scale sampling strategy.} 
After the exploration of codeword selection, we investigate the performance of our proposed multi－scale dense sampling strategy for patch extraction. In this exploration study, we choose four types of encoding methods: (1) average pooling over patch descriptors $\mathbf{f}$, (2) Fisher vector, (3) VLAD, and (4) our proposed VSAD. We sample image patches from 1 scale to 9 scales, resulting in the number of patches from 200 to 1800. The experimental results are summarized in the middle of Figure \ref{fig:exploration}. We notice that the performance of traditional encoding methods (i.e., Fisher vector, VLAD) is more sensitive to the number of sampled patches, while the performance of our proposed VSAD increases gradually as more patches are sampled. We analyze that the traditional encoding methods heavily rely on unsupervised dictionary learning (i.e., GMMs, $k$-means), whose training is unstable when the number of sampled patches is small. Moreover, we observe that our VSAD representation is still able to obtain high performance when only 200 patches are sampled, which again demonstrates the effectiveness of semantic codebook and VSAD representations. For real application, we may simply sample 200 patches for fast processing, but to fully reveal the representation capacity of VSAD,  we crop image patches from 9 scales in the remaining experiments.

\textbf{Exploration on dimension reduction.}
The dimension of scene-PatchNet descriptor $\mathbf{f}$ is relatively high (1,024) and it may be possible to reduce its dimension for VSAD representation. So we perform experiments to study the effect of dimension reduction on scene-PatchNet descriptor. The numerical results are reported in the right of Figure \ref{fig:exploration} and the performance difference is relatively small for different dimensions (the maximum performance difference is around 0.5\%). We also see that PCA dimension reduction can not bring the performance improvement for VSAD representation, which is different from traditional encoding methods (e.g., Fisher vector, VLAD). This result could be explained by two possible reasons: (1) PatchNet descriptors are more discriminative and compact than hand-crafted features and dimension reduction may cause more information loss; (2) Our VSAD representation is based on the semantic codebook, which does not rely on any unsupervised learning methods (e.g., GMMs, $k$-means). Therefore de-correlating different dimensions of descriptors can not bring any advantage for semantic dictionary learning. Overall, in the case of fully exploiting the representation power of VSAD, we could keep the dimension of PatchNet descriptor as 1,024, and in the case of high computational efficiency, we could choose the dimension as 100 for fast processing speed and low dimensional representation.

\begin{table}[t]
\caption{Comparison of different structures for the PatchNet design on the dataset of MIT Indoor67.}
\label{table:structure}
\resizebox{\linewidth}{!}{
\begin{tabular}{|l|c|}
\hline
Descriptor $\mathbf{f}$ & MIT Indoor67  \\
\hline
scene-PatchNet (VGGNet16)+ average pooling & 81.1 \\
scene-PatchNet (Inception V2) + average pooling & 78.5 \\
\hline
scene-PatchNet (VGGNet16)+ VLAD & 83.7 \\
scene-PatchNet (Inception V2) + VLAD & 83.9 \\
\hline
scene-PatchNet (VGGNet16)+ Fisher vector & 81.2 \\
scene-PatchNet (Inception V2) + Fisher vector & 83.6 \\
\hline
scene-PatchNet (VGGNet16)+ VSAD & 83.9 \\
scene-PatchNet (Inception V2) + VSAD & 84.9 \\
\hline
\end{tabular}
}
\end{table}

\textbf{Exploration on network architectures}
We explore different network architectures to verify the effectiveness of PatchNet and VSAD representation on the MIT Indoor67 dataset. Specifically, we compare two network structures: VGGNet16 and Inception V2. The implementation details of VGGNet16 PatchNet are the same with those of Inception V2 PatchNet, as described in Section \ref{sec:detail}. We also train two kinds of PatchNets for VGGNet16 structure, namely object-PatchNet on the ImageNet dataset and scene-PatchNet on the Places dataset, where the top5 classification accuracy is 80.1\% and 82.9\%, respectively. As the last hidden layer (fc7) of VGGNet16 has a much higher dimension (4096), we decreases its dimension to 100 as the patch descriptor $\mathbf{f}$ for computational efficiency. For patch aggregating, we use the semantic probability from object-PatchNet, where we select the most 256 discriminative object classes. The experimental results are summarized in Table~\ref{table:structure} and two conclusions can be drawn from this comparison. First, for both structures of VGGNet16 and Inception V2, our VSAD representation outperforms other three encoding methods. Second, the recognition accuracy of Inception V2 PatchNet is slightly better than that of VGGNet16 PatchNet, for all aggregation based encoding methods, including VLAD, Fisher vector, and VSAD. So, in the following experiment, we choose the Inception V2 as our PatchNet structure.

\subsection{Evaluation on PatchNet architectures}
\begin{table}[t]
\caption{Comparison of PatchNet and ImageCNN for patch modeling on the dataset of MIT Indoor67.}
\label{table:feature_p_exploration}
\resizebox{\linewidth}{!}{
\begin{tabular}{|l|c|c|}
\hline
Descriptor $\mathbf{f}$ & object-PatchNet $\mathbf{p}$ & object-ImageCNN $\mathbf{p}$ \\
\hline
scene-PatchNet (1,024D) & 84.9 & 84.7\\
scene-PatchNet (100D) & 84.3 & 84.0\\
scene-ImageCNN (1,024D) & 83.8 & 83.4\\
scene-ImageCNN (100D) & 83.6 & 83.1\\
\hline
objcet-PatchNet (1,024D) & 79.6 & 79.4\\
object-PatchNet (100D) & 79.5 & 79.3\\
object-ImageCNN (1,024D) & 79.3 & 79.2\\
object-ImageCNN (100D) & 79.1 & 78.7\\
\hline
\end{tabular}
}
\end{table}

After exploring the important parameters of our method, we focus on verifying the effectiveness of PatchNet on patch modeling in this subsection. Our PatchNet is a patch-level architecture, whose hidden layer activation features $\mathbf{f}$ could be exploited to describe patch appearance and prediction probability $\mathbf{p}$ to aggregate these patches. In this subsection we compare two network architectures: image-level CNNs (ImageCNNs) and patch-level CNNs (PatchNets), and demonstrate the superior performance of PatchNet on describing and aggregating local patches on the dataset of MIT Indoor67. 

For fair comparison, we also choose the Inception V2 architecture \cite{IoffeS15} as our ImageCNN structure, and following the similar training procedure to PatchNet, we learn the network weights on the datasets of ImageNet~\cite{DengDSLL009} and Places~\cite{ZhouLXTO14}. The resulted CNNs are denoted as {\bf object-ImageCNN} and {\bf scene-ImageCNN}. The main difference between PatchNet and ImageCNN is their receptive filed, where PatchNet operates on the local patches ($128 \times 128$), while ImageCNN takes the whole image ($224 \times 224$) as input. In this exploration experiment, we investigate four kinds of descriptors, including $\mathbf{f}$ extracted from scene-PatchNet, scene-ImageCNN, object-PatchNet, and object-ImageCNN. Meanwhile, we compare the descriptor $\mathbf{f}$ without dimension reduction (i.e., 1,024) and with dimension reduction to 100. For aggregating semantic probability $\mathbf{p}$, we choose two types of probabilities from object-PatchNet and object-ImageCNN respectively.

The experiment results are summarized in Table~\ref{table:feature_p_exploration} and several conclusions can be drawn as follows: (1) From the comparison between object network descriptors and scene network descriptors, we see that scene network descriptor is more suitable for recognizing the categories from MIT Indoor67, no matter which architecture and aggregating probability is chosen; (2) From the comparison between descriptors from image-level and patch-level architectures, we conclude that PatchNet is better than ImageCNN. This superior performance of descriptors from PatchNet indicates the effectiveness of training PatchNet for local patch description; (3) From the comparison between aggregating probabilities from PatchNet and ImageCNN, our proposed PatchNet architecture again outperforms the traditional image-level CNN, which implies the semantic probability from the PatchNet is more suitable for VSAD representation. Overall, we empirically demonstrate that our proposed PatchNet architecture is more effective for describing and aggregating local patches.

\subsection{Evaluation on Aggregating Patches}

\begin{table}[t]
\caption{Performance comparison with SIFT descriptors on the datasets of MIT Indoor67 and SUN397.}
\label{table:sift_result}
\resizebox{\linewidth}{!}{
\begin{tabular}{|l|c|c|}
\hline
Method & MIT indoor67 & SUN397\\
\hline\hline
SIFT+VLAD & 32.6 & 19.2\\
SIFT+FV & 42.8 & 24.4\\
Dense-Multiscale-SIFT+VLAD+aug. \cite{vedaldi08vlfeat} & 53.3 & - \\
Dense-Multiscale-SIFT+Fisher vector \cite{vedaldi08vlfeat} & 58.3 & - \\
Dense-Multiscale-SIFT+Fisher vector \cite{XiaoHEOT10} & - & 38.0 \\
\hline
SIFT+ VSAD & \textbf{60.8} & \textbf{40.3}\\
\hline
\end{tabular}
}
\end{table}

\begin{table}[t]
\caption{Performance Comparison with scene-PatchNet descriptor on the datasets of MIT Indoor67 and SUN397.}
\begin{center}
\begin{tabular}{|l|c|c|}
\hline
Method & MIT indoor67 & SUN397\\
\hline\hline
scene-PatchNet+average pooling & 78.5 & 63.5\\
scene-PatchNet+Fisher vector & 83.6 & 69.0\\
scene-PatchNet+VLAD & 83.9 & 70.1\\
\hline
scene-PatchNet+VSAD & \textbf{84.9} & \textbf{71.7}\\
\hline
\end{tabular}
\end{center}
\label{table:scene_patchnet_result}
\end{table}

\begin{table}[t]
\caption{Performance Comparison with concatenated descriptor (Hybrid-PatchNet) from object-PatchNet and scene-PatchNet on the datasets of MIT Indoor67 and SUN397.}
\begin{center}
\begin{tabular}{|l|c|c|}
\hline
Method & MIT indoor67 & SUN397\\
\hline\hline
hybrid-PatchNet+average pooling & 80.6 & 65.7\\
hybrid-PatchNet+Fisher vector & 82.6 & 68.4 \\
hybrid-PatchNet+VLAD & 84.9 & 70.9 \\
\hline
hybrid-PatchNet+VSAD & \textbf{86.1} & \textbf{72.0} \\
\hline
\end{tabular}
\end{center}
\label{table:hybrid_patchnet_result}
\end{table}

In this subsection we focus on studying the effectiveness of PatchNet on aggregating local patches. We perform experiments with different types of descriptors and compare VSAD with other aggregation based encoding methods, including average pooling, Fisher vector (FV), and VLAD, on both datasets of MIT Indoor67 and SUN397. 

\textbf{Performance with SIFT descriptors.}
We first verify the effectiveness of our VSAD representation by using the hand-crafted features (i.e., SIFT \cite{Lowe04}). For each image, we extract the SIFT descriptors from image patches (in grid of $64\times 64$, a stride of 16 pixels). These SIFT descriptors are square-rooted and then de-correlated by PCA processing, where the dimension is reduced from 128 to 80. We compare our VSAD with traditional encoding methods of VLAD \cite{JegouPDSPS12} and Fisher vector \cite{SanchezPMV13}. For traditional encoding methods, we directly learn the codebooks with unsupervised learning methods (i.e., GMMs, $k$-means) based on SIFT descriptors, where the codebook size is set as 256. For our VSAD, we first resize the extracted patches of training images to $128 \times 128$. Then we feed them to the learned object-PatchNet and obtain their corresponding semantic probabilities $\mathbf{p}$. Based on the SIFT descriptors $\mathbf{f}$ and the semantic probabilities $\mathbf{p}$ of these training patches, we construct our semantic codebook and VSAD representations by Equation (\ref{equ:mean}) and (\ref{equ:var_vsad}). 

The experimental results are reported in Table \ref{table:sift_result}. We see that our VSAD significantly outperforms the traditional VLAD and Fisher vector methods on both datasets of MIT Indoor67 and SUN397. Meanwhile, we also list the performance of VLAD and Fisher vector with multi-scale sampled SIFT descriptors from previous works~\cite{vedaldi08vlfeat,XiaoHEOT10}. Our VSAD from single-scale sampled patches is still better than the performance of traditional methods with multi-scale sampled patches, which demonstrates the advantages of semantic codebook and VSAD representations.

\textbf{Performance with scene-PatchNet descriptors.}
After evaluating VSAD representation with SIFT descriptors, we are ready to demonstrate the effectiveness of our complete framework, i.e. \emph{describing} and \emph{aggregating} local patches with PatchNet. According to previous study, we choose the multi-scale dense sampling method (9 scales) to extract patches. For each patch, we extract the scene-PatchNet descriptor $\mathbf{f}$ and use the semantic probabilities $\mathbf{p}$ obtained from object-PatchNet to aggregate these descriptors.

\begin{figure*}
\centering
	\includegraphics[width=0.75\textwidth]{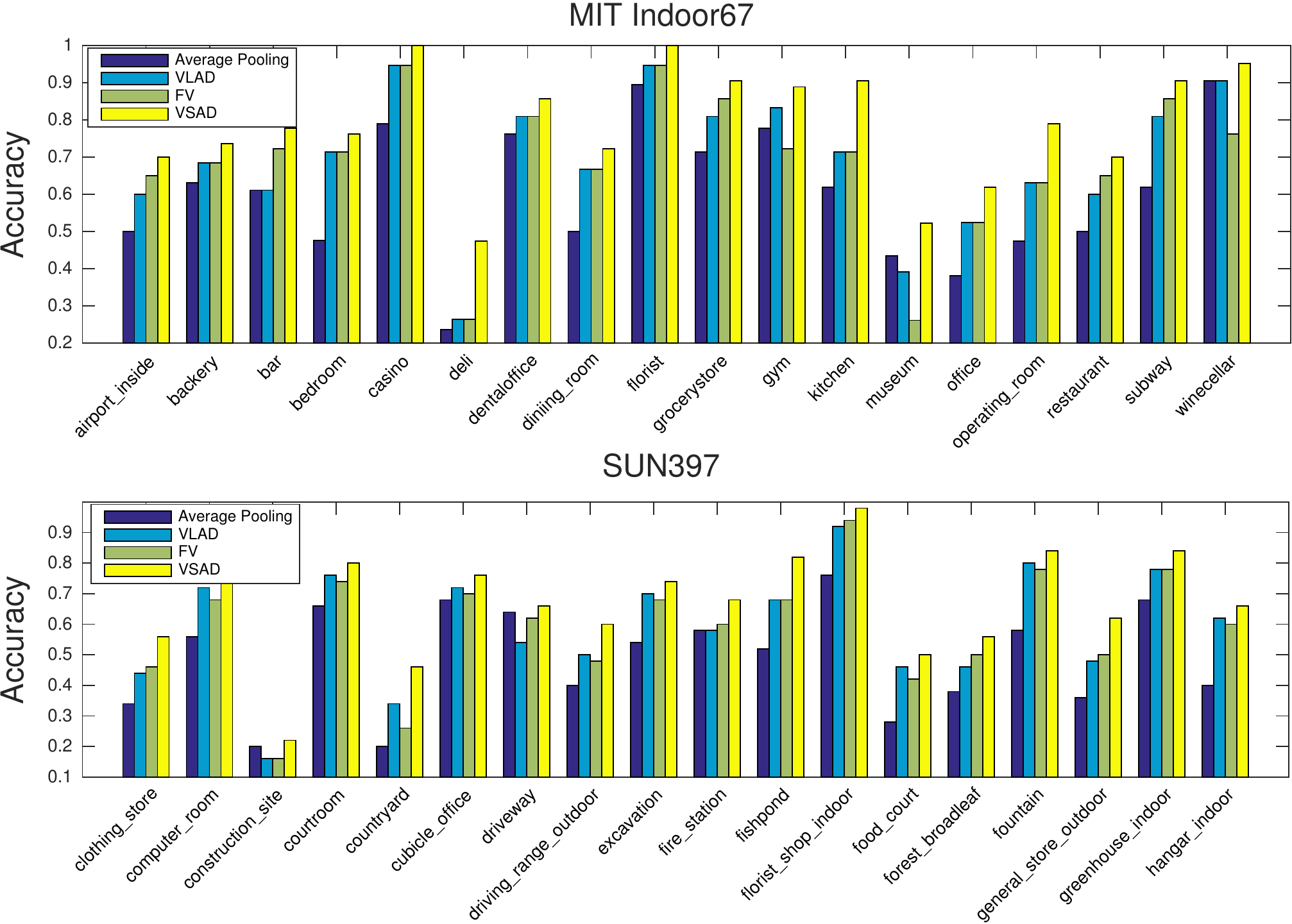}
	\caption{Several categories with significant improvement on MIT Indoor67 and SUN397. These results show the strong ability of VSAD encoding for scene recognition.}
    \label{fig:mit_sun_comparison}
    \vspace{-5mm}
\end{figure*}

We make comparison among the performance of VSAD, Average Pooling, Fisher vector, and VLAD. For fair comparison, we fix the dimension of PatchNet descriptor as 1,024 for all encoding methods, but de-correlate different dimensions to make GMM training easier. The numerical results are summarized in Table \ref{table:scene_patchnet_result} and our VSAD encoding method achieves the best accuracy on both datasets of MIT Indoor67 and SUN397. Some more detailed results are depicted in Figure \ref{fig:mit_sun_comparison}, where we show the classification accuracy on a number of scene categories from the MIT Indoor67 and SUN397. VSAD achieves a clear performance improvement over other encoding methods.

\textbf{Performance with hybrid-PatchNet descriptors.}
Finally, to further boost the performance of VSAD representation and make comparison more fair, we extract two descriptors for each patch, namely descriptor from scene-PatchNet and object-PatchNet. We denote this fused descriptor as hybrid-PatchNet descriptor. For computational efficiency, we first decrease the dimension of each descriptor to 100 for feature encoding. Then, we concatenate the image-level representation from two descriptors as the final representation. As shown in Table~\ref{table:hybrid_patchnet_result}, our VSAD encoding still outperforms other encoding methods, including average pooling, VLAD, Fisher vector, with this new hybrid-PatchNet descriptor, which further demonstrates the effectiveness of PatchNet for describing and aggregating local patches.

\begin{table}[t]
\caption{Comparison with Related Works on MIT Indoor67. Note that the codebook of FV and our VSAD are encoded by deep feature from our scene-PatchNet.}
\vspace{-3mm}
\begin{center}
\begin{tabular}{l|c|c}
\hline
\textbf{Method} & \textbf{Publication} & \textbf{Accuracy}(\%)\\
\hline\hline
Patches+Gist+SP+DPM \cite{UDMDP} & ECCV2012 & 49.4\\
BFO+HOG \cite{Bfo} & CVPR2013 & 58.9\\
FV+BoP \cite{JunejaVJZ13} & CVPR2013 & 63.1\\
FV+PC \cite{NIPS2013} & NIPS2013 & 68.9\\
FV(SPM+OPM) \cite{OPM} & CVPR2014 & 63.5\\
Zhang \emph{et al.} \cite{TIP14} & TIP2014 & 39.9\\
DSFL \cite{zuozhen} & ECCV2014 & 52.2\\
LCCD+SIFT \cite{arXiv:1508.00307} & arXiv2015 & 66.0\\
\hline
OverFeat+SVM \cite{RazavianASC14} & CVPRW2014 & 69.0\\
AlexNet fc+VLAD\cite{GongWGL14} & ECCV2014 & 68.9\\
DSFL+DeCaf \cite{zuozhen} & ECCV2014 & 76.2\\
DeCaf \cite{DonahueJVHZTD14} & ICML2014 & 59.5\\
DAG+VGG19 \cite{dag} & ICCV15 & 77.5\\
C-HLSTM \cite{arXiv:1509.03877} & arXiv2015 & 75.7\\
VGG19 conv5+FV \cite{GaoWWL15} & arXiv2015 & 78.3\\
Places205-VGGNet-16	\cite{WangGHQ15} & arXiv2015 & 81.2\\
VGG19 conv5+FV \cite{CimpoiMKV15} & CVPR2015 & 81.0\\
Semantic FV \cite{DixitCGRV15} & CVPR2015 & 72.9\\
LS-DHM \cite{arXiv:1601.07576} & TIP2017 & 83.8\\
%Hybrid CNN \cite{arXiv:1601.07977} & arXiv2016 & \textbf{85.9}\\
%Herranz \emph{et al.} \cite{cvpr16_scene} & CVPR2016 & \textbf{86.0}\\
\hline
%Our VSAD & - & 79.6\\
Our VSAD & - & 84.9\\
%Our VSAD+FV & - & 79.9\\
Our VSAD+FV & - & 84.4\\
%Our VSAD+Places205-VGGNet-16 & - & 84.8\\
Our VSAD+Places205-VGGNet-16 & - & 85.3\\
%Our VSAD+FV+ Places205-VGGNet-16 & - & \textbf{84.9}\\
Our VSAD+FV+ Places205-VGGNet-16 & - & \textbf{86.2}\\
\hline
\end{tabular}
\end{center}
\label{table:relate_mit}
\vspace{-5mm}
\end{table}

\begin{table}[t]
\caption{Comparison with Related Works on SUN397. Note that the codebook of FV and our VSAD are encoded by deep feature from our PatchNet. Our VSAD in combination with Places205-VGGNet-16 outperform state-of-the-art and surpass human performance.}
\vspace{-3mm}
\begin{center}
\begin{tabular}{l|c|c}
\hline
\textbf{Method} & \textbf{Publication} & \textbf{Accuracy}(\%)\\
\hline\hline
Xiao \emph{et al.} \cite{XiaoHEOT10} & CVPR2010 & 38.0\\
\hline
FV(SIFT+LCS) \cite{SanchezPMV13} & IJCV2013 & 47.2\\
FV(SPM+OPM) \cite{OPM} & CVPR2014 & 45.9\\
LCCD+SIFT \cite{arXiv:1508.00307} & arXiv2015 & 49.7\\
\hline
DeCaf \cite{DonahueJVHZTD14} & ICML2014 & 43.8 \\
AlexNet fc+VLAD \cite{GongWGL14} & ECCV2014 & 52.0\\
Places-CNN \cite{ZhouLXTO14} & NIPS2014 & 54.3\\
Semantic FV \cite{DixitCGRV15} & CVPR2015 & 54.4\\
VGG19 conv5+FV \cite{GaoWWL15} & arXiv2015 & 59.8\\
Places205-VGGNet-16	\cite{WangGHQ15}  & arXiv2015 & 66.9\\
LS-DHM \cite{arXiv:1601.07576} & TIP2017 & 67.6\\
%Herranz \emph{et al.} \cite{cvpr16_scene} & CVPR2016 & 70.2\\
\hline
%Our VSAD & - & 63.6\\
%Our VSAD+FV & - & 64.2\\
\textbf{Human performance} \cite{XiaoHEOT10} & CVPR2010 & \textbf{68.5} \\
Our VSAD & - & 71.7\\
Our VSAD+FV & - & 72.2\\
%\textbf{Human performance} \cite{XiaoHEOT10} & CVPR2010 & \textbf{68.5} \\
Our VSAD+Places205-VGGNet-16  & - & \textbf{72.5} \\
Our VSAD+FV+ Places205-VGGNet-16 & - & \textbf{73.0} \\
\hline
\end{tabular}
\end{center}
\label{table:relate_sun}
\vspace{-5mm}
\end{table}

\subsection{Comparison with the State of the Art}

After the exploration of different components of our proposed framework, we are ready to present our final scene recognition method in this subsection and compare its performance with these sate-of-the-art methods. In our final recognition method, we choose the VSAD representations by using scene-PatchNet to describe each patch ($\mathbf{f}$) and object-PatchNet to aggregate these local pathces ($\mathbf{p}$). Furthermore, we combine our VSAD representation, with Fisher vector and deep features of Place205-VGGNet-16~\cite{WangGHQ15} to study the complementarity between them, and achieve the new state of the art on these two challenging scene recognition benchmarks.

The results are summarized in Table \ref{table:relate_mit} and Table \ref{table:relate_sun}, which show that our VSAD representation outperforms the previous state-of-the-art method (LS-DHM~\cite{arXiv:1601.07576}). Furthermore, we explore the complementary properties of our VSAD from the following three perspectives.
(1) The semantic codebook of our VSAD is generated by our discriminative PatchNet, while the traditional codebook of Fisher vector (or VLAD) is generated in a generative and unsupervised manner.
Hence, we combine our VSAD with Fisher vector to integrate both discriminative and generative power.
As shown in Table \ref{table:relate_mit} and Table \ref{table:relate_sun}, the performance of this combination further improves the accuracy.
(2) Our VSAD is based on local patches and is complementary to those global representations of image-level CNN. Hence, we combine our VSAD and the deep global feature (in the FC6 layer) of Place205-VGGNet-16~\cite{WangGHQ15} to take advantage of both patch-level and image-level features.
The results in Table \ref{table:relate_mit} and Table \ref{table:relate_sun} show that this combination surpasses the human performance on SUN 397 dataset.
(3) Finally,
we combine our VSAD, Fisher vector, and deep global feature of Place205-VGGNet-16 to put the state-of-the-art performance forward with a large margin.
To our best knowledge, the result of this combination in Table \ref{table:relate_mit} and Table \ref{table:relate_sun} is one of the best performance on both MIT Indoor67 and SUN397, which surpasses human performance $(68.5\%)$ on SUN 397 by 4 percents.

\begin{figure*}
    \includegraphics[width=\textwidth]{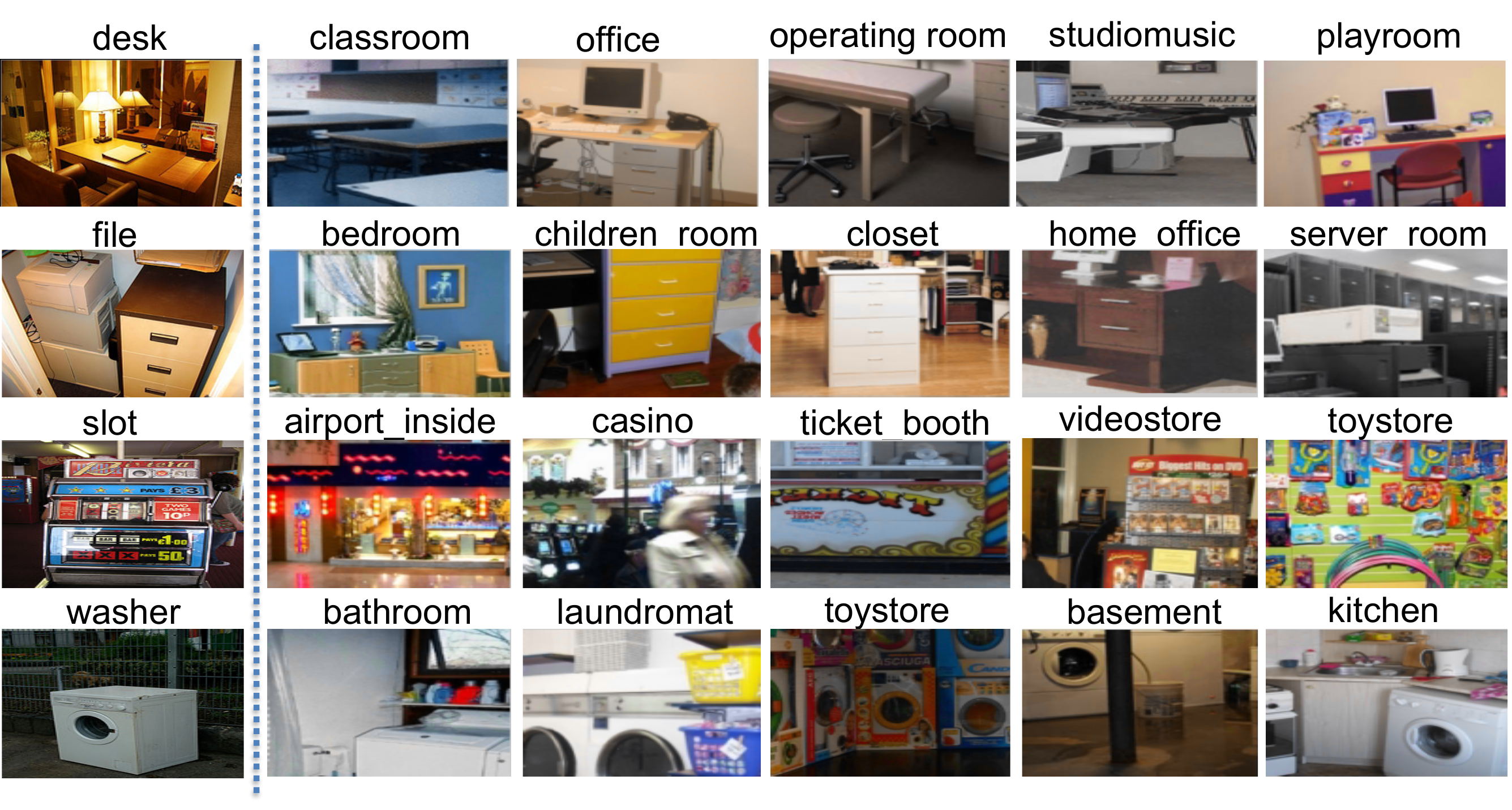}
	\caption{{\bf Analysis of semantic codebook}. The codeword (the 1st column) appears in its related scene categories (the 2nd-5th column), which illustrates that our codebook contains important semantic information.}
    \label{fig:codeword_scenepatch}
    \vspace{-5mm}
\end{figure*}
\subsection{Visualization of Semantic Codebook}

Finally, we show the importance of object-based semantic codebook in Figure \ref{fig:codeword_scenepatch}.
Here we use four objects from ImageNet (\emph{desk}, \emph{file}, \emph{slot}, \emph{washer}) as an illustration of the codewords in our semantic codebook.
For each codeword,
we find five scene categories from either MIT Indoor67 or SUN 397 (the 2nd to 5th column of Figure \ref{fig:codeword_scenepatch}),
based on their semantic conditional probability (more than 0.9) with respect to this codeword.
As shown in Figure \ref{fig:codeword_scenepatch},
the object (codeword) appears in its related scene categories, which makes our codebook contains important semantic cues to improve the performance of scene recognition.

\section{Conclusions}
\label{sec:con}

In this paper we have designed a patch-level architecture to model local patches, called as {\em PatchNet}, which is trainable in an end-to-end manner with a weakly supervised setting. To fully unleash the potential of PatchNet, we proposed a hybrid visual representation, named as {\em VSAD}, by exploiting PatchNet to both describe and aggregate these local patches, whose superior performance was verified on two challenging scene benchmarks: MIT indoor67 and SUN397. The excellent performance demonstrates the effectiveness of PatchNet for patch description and aggregation.

\bibliographystyle{IEEEtran}
\bibliography{egbib}

\end{document}